\documentclass[conference]{IEEEtran}
\usepackage{cite}
\usepackage{amsmath,amssymb,amsfonts}
\usepackage{graphicx}
\usepackage{textcomp}
\usepackage{xcolor}
\usepackage{fancyhdr}
\usepackage[hyphens]{url}

\def\BibTeX{{\rm B\kern-.05em{\sc i\kern-.025em b}\kern-.08em
    T\kern-.1667em\lower.7ex\hbox{E}\kern-.125emX}}

\fancypagestyle{firstpage}{
  \fancyhf{}

  \fancyhead[C]{\normalsize{ML for Computer Architecture and Systems 2021}} 
  \fancyfoot[C]{\thepage}
}

\newcommand{\eps}{\varepsilon}

\DeclareMathOperator{\EE}{\mathbb{E}}
\newcommand{\Prob}[1]{\ensuremath{\mathbb{P}\left(#1\right)}}
\newcommand{\expect}[1]{\ensuremath{\operatorname{\EE}\left[#1\right]}}

\newcommand{\pigreedy}[1]{\ensuremath{\operatorname{\pi_{\footnotesize \text{greedy}}}\left(#1\right)}}

\newcommand{\states}{\mathcal{S}}
\newcommand{\actions}{\mathcal{A}}
\newcommand{\rewards}{\mathcal{R}}

\newcommand{\RR}{\mathbb{R}}

\newcommand{\e}{\ensuremath{{\textrm e}}}

\newcommand{\TAKEN}{\text{T}}
\newcommand{\NOTTAKEN}{\text{NT}}

\newcommand{\x}{\bm{x}}
\newcommand{\q}{\bm{q}}
\newcommand{\vectheta}{\bm{\theta}}
\newcommand{\zero}{\bm{0}}

\usepackage{csquotes} 

\usepackage{algorithm}
\usepackage{algpseudocode}
\usepackage{bm} 
\usepackage{booktabs}
\usepackage{listings}
\usepackage{color}
\usepackage[normalem]{ulem}

\definecolor{mygreen}{rgb}{0,0.6,0}
\definecolor{mygray}{rgb}{0.5,0.5,0.5}
\definecolor{mymauve}{rgb}{0.58,0,0.82}

\usepackage{soul,color}
\soulregister\cite7
\soulregister\ref7
\soulregister\textsubscript7
\soulregister{\etal}{7}

\newcommand{\GQlacron}{G-QLAg}
\newcommand{\GQlmethod}{Global-History $Q$-Learning Agent}
\soulregister\GQlacron{1}
\soulregister\GQlmethod{1}

\newcommand{\Pgacron}{PolGAg}
\newcommand{\Pgmethod}{Policy Gradient Agent}
\soulregister\Pgacron{1}
\soulregister\Pgmethod{1}

\newcommand{\gymBP}{\textit{gym\_bp\_cbp2016}}

\usepackage{xcolor}
\ifdefined\RELEASEON
    \newcommand\boris[1]{}
    \newcommand\tassos[1]{}
    \newcommand\kleov[1]{}
    
\else
    \newcommand\boris[1]{{\color{blue}[Boris]: #1}}
    \newcommand\tassos[1]{{\color{orange}[Tassos]: #1}}
    \newcommand\kleov[1]{{\color{red}[Kleov]: #1}}
\fi

\title{Branch Prediction as a Reinforcement Learning Problem: Why, How and Case Studies}
\author{\IEEEauthorblockN{Anastasios Zouzias}
\IEEEauthorblockA{Huawei Technologies\\
Zurich Research Center\\
Switzerland\\
anastasios.zouzias@huawei.com}
\and
\IEEEauthorblockN{Kleovoulos Kalaitzidis}
\IEEEauthorblockA{Huawei Technologies\\
Zurich Research Center\\
Switzerland\\
kleovoulos.kalaitzidis@huawei.com}
\and
\IEEEauthorblockN{Boris Grot}
\IEEEauthorblockA{University of Edinburgh\\
School of Informatics\\
United Kingdom\\
boris.grot@ed.ac.uk}}

\begin{document}
\maketitle
\thispagestyle{firstpage}
\pagestyle{plain}

\begin{abstract}
Recent years have seen stagnating improvements to branch predictor (BP) efficacy and a dearth of fresh ideas in branch predictor design, calling for fresh thinking in this area. This paper argues that looking at BP from the viewpoint of Reinforcement Learning (RL) facilitates systematic reasoning about, and exploration of, BP designs. We describe how to apply the RL formulation to branch predictors, show that existing predictors can be succinctly expressed in this formulation, and study two RL-based variants of conventional BPs.
\end{abstract}
%
\section{Introduction}\label{sec:intro}
%
Branch prediction (BP) lies at the heart of high-performance processor design as it enables larger instruction windows that are imperative for extracting instruction- and memory-level parallelism. Today's state-of-the-art branch predictors learn correlations between branches through the use of large hardware tables, with efficacy strongly correlated with the amount of storage available. However, with the looming end of Moore's law, processor designers are faced with the challenge of improving performance, including that of branch predictors, without the benefit of larger transistor budgets. 

Today, virtually all branch predictors in high-end processors are highly-engineered variants of TAGE~\cite{seznec2006} and Perceptron~\cite{perceptron2001}.
Despite extensive differences in the prediction mechanisms of these two families, 
predictors in each of these families are produced through meticulous feature engineering, often driven by a combination of intuition and simulator-driven ``tweaking".
Perhaps not surprisingly, improvements to both families of BPs have stagnated in recent years, and the question facing the entire processor industry is how to uncover new optimizations and entirely new designs that can outperform today's best.

This paper observes that branch prediction can be expressed as a {\em Reinforcement Learning (RL)} problem. In RL, the learning agent seeks to find a (near-)optimal policy that maximises the reward function by interacting, and learning from, the environment~\cite{book:Sutton1998}. Viewed through this lense,  a branch predictor is an agent that observes the program's control flow (i.e., the history of branch outcomes) and tries to learn a policy that maximizes the accuracy of future control-flow predictions.
We argue that viewing BP as an RL problem enables a systematic approach to model and explore branch predictor designs through explicit reasoning of each aspect of the predictor, such as its state representation, decision-making policy, and strategy to minimize the number of mispredictions. We showcase how existing predictors seamlessly map to the RL formulation, and present two RL-based variants of predictor designs. Finally, we highlight new promising research venues that we believe RL opens for BP.

\section{Reinforcement Learning Background} \label{sec:rl-basics}
%
Reinforcement learning algorithms are well-suited for scenarios where an \emph{agent} can learn from an \emph{environment}. The agent interacts with the environment using a set of actions. After the agent selects an action, the environment responds with the action's reward and the next environment state. The state set and state transitions of the environment are commonly modelled with a \emph{Markov Decision Process}\footnote{A \emph{Markov Decision Process} is a tuple ($\states$,$\actions$,P,$\rewards$, $\gamma$), where $\states$ is a set of states, $\actions$ is a finite set of actions, $P$ is the state transition probability matrix $P_{ss'}^{a}= \Prob{S_{t+1}=s' | S_t =s, A_t = a}$, $\rewards$ is a reward function $\rewards:=\expect{R_{t+1} | S_t = s, A_t = a}$ and $\gamma\in{[0,1)}$ is a discount factor where $\expect{\cdot | \cdot}$ denotes the conditional expectation.} (MDP)~\cite{book:Sutton1998}. The goal of an agent is to take actions that maximize the future cumulative reward $G_t$ for any time-step $t$, i.e., $G_t:= R_{t+1}+\gamma R_{t+2}+\dots = \sum_{k=0}^{\infty} \gamma^{k}R_{t+k+1}$.

%
RL agents are objective-driven and characterized by their \emph{policy} (behavior) on any given state. A policy $\pi$ is a probability distribution over actions given a state, i.e., $\pi(a|s):= \Prob{A_t = a | S_t = s}$. The state-action value function $Q_{\pi}(s, a)$ (known as $Q$-values) is the expected $G_t$ starting from $s$, taking an action $a$, and then following the policy $\pi$, i.e.,
%
\begin{equation}\label{eqn:state_action_value_fnc}
    Q_{\pi}(s, a):= \EE_{\pi}\left[G_t | S_t = s, A_t = a \right].
\end{equation}
%
Given the state-action value function, the \emph{greedy} policy is defined as $\pigreedy{s}:=\arg\max_{a\in\actions}{Q(s,a)}$. Similarly, for $0\leq \eps<1$, the \emph{$\eps$-greedy} policy is a policy that equals $\pigreedy{s}$ with probability $1-\eps$ and selects an action uniformly at random with probability $\eps$.
%

%
Depending on whether the agent models the environment (i.e., models the reward function and/or transition probabilities), RL methods are categorized in \emph{model-based} and \emph{model-free}. In this work, we focus\footnote{Model-free aim to \emph{only} learn an optimal policy, i.e., similarly to branch prediction that aims for a specific speculation policy, whereas model-based methods aim to also model the environment.} on the model-free category in which there are two main types of RL methods: {\em tabular} (Section~\ref{sec:rl:tabular}) and {\em function approximation} (Section~\ref{sec:rl:policy-gradient-intro}). In tabular methods, the agent tries to learn from the environment by storing/updating information in fixed-sized tables where each entry corresponds to a state/action pair. In function approximation methods, the agent directly optimizes a parametrized function of the policy.
%
%
\vspace{-2mm}
\subsection{Tabular Methods: Q-Learning}\label{sec:rl:tabular}
%
A number of tabular RL methods exist; most popular ones include TD-learning~\cite{book:Sutton1998}, SARSA~\cite{rl:sarsa}, Q-Learning~\cite{rl:qlearning} and double Q-Learning~\cite{rl:double-qlearning}. Here we focus on the $Q$-Learning algorithm that provides specific convergence guarantees~\cite{rl:qlearning}\footnote{We have experimented with most of the tabular methods and found their branch prediction performance to be similar.}. 
$Q$-Learning stores the $Q$-values $Q(s,a)$ for every state and action pair in a fixed-sized table. Given a state $s$ from the environment, $Q$-Learning predicts the action greedily using the policy $\pigreedy{s}$. The $Q$-Learning update rule works as follows: (a) choose an $a\in \actions$ based on the current state using the $\eps$-greedy policy; (b) observe reward $r$, next state $s'$ after action $a$; (c) update $Q(s,a) \mathrel{{+}{=}} \alpha [r+\gamma \max_{a'\in{\actions}} Q(s', a') - Q(s,a)]$ for a positive learning rate parameter $\alpha$.
%
\vspace{-1mm}
\subsection{Function Approximation Methods: Policy Gradient}\label{sec:rl:policy-gradient-intro}
%
In function approximation methods, the policy is parametrized using a vector parameter $\vectheta$ and is denoted by $\pi_{\vectheta}(a|s)$. The goal is to optimize the expected cumulative reward by maximizing
%
\begin{equation}\label{eqn:pg_objective}
    J(\vectheta) := \sum_{s\in{\states}} d^{\pi}(s) \sum_{a\in\actions} \pi_{\vectheta}(a|s) Q_{\pi_{\vectheta}}(s, a),
\end{equation}
%
where $d^{\pi}(s)$ is the stationary distribution\footnote{In general, a branch predictor might never converge to a stationary distribution due to the partial-observability of the state space. As an example, consider data-dependent branches whereby branch outcomes are correlated with data values but the actual data values are not visible to the branch predictor.} of the Markov Chain defined by $\pi_{\vectheta}$ on a given MDP. Using gradient ascent, $\vectheta$ can be updated towards the direction of the gradient $\nabla J(\vectheta)$ to find an (local) optimal $\vectheta$ that maximizes the average future cumulative reward, i.e., $\vectheta \leftarrow \vectheta + \alpha \nabla J(\vectheta)$.
As a representative example, the REINFORCE Policy-Gradient method~\cite{book:Sutton1998,rl:REINFORCE}
works as follows:
%
\begin{itemize}
    \item Collect reward $r$ on state $s\in\states$ with action $a\in\actions$.
    \item Update policy by maximizing $J(\vectheta)$ via an optimization strategy (i.e., online gradient ascent).
\end{itemize}
%
Given a discrete action space (for example of size two), the policy is usually parametrized by a softmax function $\pi_{\vectheta}(a|s) := \e^{h(s,a,\vectheta)}/  \sum_{a\in{\actions}} \e^{h(s,a,\vectheta)}$ for an arbitrary differentiable function $h(s,a,\vectheta)$.
The second step of the Policy-Gradient method requires a gradient computation. In Section~\ref{sec:PolGAg}, a special case will be considered where $h$ is a linear function on $\vectheta$, i.e., $h(s,a,\vectheta) := \vectheta^\top \x(s,a)$, where  $\x(s,a)$ is a vector representation of the state-action space and $\vectheta^\top \x$ denotes vector dot-product. In this specific case, the policy gradient theorem~\cite[Chapter~13]{book:Sutton1998} gives us a closed form solution for the gradient, i.e., $\nabla J(\vectheta) = \x(s,a) - \bm{\mu}$, where $\bm{\mu} := \sum_{a\in\actions} \pi_{\vectheta}(a|s) \x(s,a)$. 
%
\section{Reinforcement Learning for Branch Prediction} \label{sec:bp_as_rl}
%
\subsection{Branch Prediction as an RL Problem}
%
%
\begin{figure}[t]
\begin{center}
  \includegraphics[width=\linewidth]{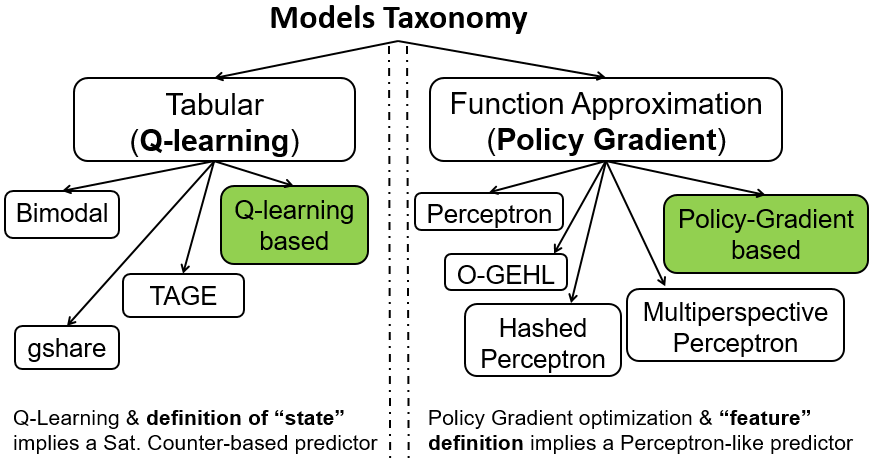}
\end{center}
\vspace{-3mm}
\caption{Taxonomy of model-free RL methods and branch predictors.}
\vspace{-3mm}
\label{fig:models-tree}
\end{figure}
%
In this section, we show that branch prediction is a task that can be seamlessly formulated as an RL problem as it abides by similar theoretical principles. 
%
%
%
%
%
In particular, through the lenses of RL, branch prediction is viewed as a pure online optimization task where the high-level goal is to minimize the rate of branch mispredictions. In RL parlance: 
\emph{``Maximization of the future cumulative reward equals to minimization of future cumulative branch mispredictions.''}
The branch predictor corresponds to the \emph{agent} and the processor's execution environment (e.g., program counter, registers, memory/cache content, etc) corresponds to the \emph{environment}. The \emph{environment} is private and communicates its state to the branch-predictor agent so that the latter can decide which action $a$ from the action space $\actions := \{\TAKEN{}, \NOTTAKEN{}\}$\footnote{\TAKEN{} stands for \emph{taken}, \NOTTAKEN{} for \emph{not-taken}.} will be followed. Depending on the correctness of the chosen action, the environment responds back with a reward $R\in{\RR}$ that guides the predictor's update (e.g., confidence counter increase or reset on a, respectively, correct or incorrect prediction). At that point, the next state of the environment is also disclosed. Generally, the state content can consist of any information available to the BP; e.g., the branch PC address, the local and/or global branch history, etc. For instance, if the global branch history is part of the state definition, its instance after a prediction (next state) will include the predicted branch outcome (note that this is the common case of speculative history update in BP).
In analogy with RL methods, branch predictors can be categorized into two separate classes: tabular and function-approximation models. Tabular predictors model their actual predictions with a set of finite-state machine (FSMs) represented by up/down saturating counters stored in storage-constrained tables. Their goal is to couple branches with the correct FSM by matching patterns in the execution history. Examples of such predictors include the Bimodal predictor~\cite{smith1981}, two-level branch predictors, such as gshare~\cite{mcfarling1993}, and, most recently, the TAGE-based~\cite{seznec2006} predictors.
Meanwhile, existing branch predictors that fall under the function-approximation models are mainly variants of the Perceptron~\cite{perceptron2001} predictor, which is based on the homonymous online learning algorithm~\cite{perceptron1962}. Perceptron learns from previous experiences by training a single-layer neural network, which stands in contrast to the memoization approach of tabular models. 
O-GEHL~\cite{ogehl_seznec2005}, Hashed Perceptron~\cite{tarjan2005}, and, most recently, Multiperspective Perceptron~\cite{jimenez2016}, are examples of this class of predictors. 
%

%
In Figure~\ref{fig:models-tree}, we classify common branch predictors into the two categories of tabular and function-approximation models. We respectively map them to their RL-based counterparts of either $Q$-Learning or Policy-Gradient origin. In essence, tabular predictors that are based on $N$-bit saturating counters can be formulated with a $Q$-Learning based scheme, while function-approximation models are a good fit for Policy-Gradient methods. We explain this mapping for $Q$-Learning based predictors in the next  section; Policy-Gradient based schemes are analyzed later in Section~\ref{sec:rl-looking-ahead}. 
%
\vspace{-1.5mm}
\subsection{An Illustrative Example: Q-Learning Based BP} \label{sec:rl-example-ql}
%
We now give a detailed example of how an existing tabular branch predictor can be cast in RL terms. For illustrative purposes, we focus on relatively simple predictors -- Bimodal and gshare (Section~\ref{sec:PolGAg} discusses a more complex Perceptron predictor). We opt not to include TAGE in our exploration as its design principles originate from the method of partial string matching used in data compression \cite{ppm1984}. However, a similar RL-based optimization of TAGE could be feasible; we leave this case for future work.
Both Bimodal and gshare can be viewed as $Q$-Learning instances; for Bimodal the state space is the set of branch PC addresses, while for gshare it is the set of branch PC addresses exclusive-ORed with (a part of) the global branch-history register (GHR). In this simple form, the RL-based versions of these predictors do not perform any exploration (see section~\ref{sec:rl:exploration}), that is, $\varepsilon = 0$. According to the description of section~\ref{sec:rl:tabular}, their update rule will be reduced to $Q(s,a) = (1-\alpha) Q(s,a) + \alpha r$, since future rewards are not considered, and thus $\gamma = 0$.
%

\begin{figure}[t]
\begin{center}
  \includegraphics[width=1.0\linewidth]{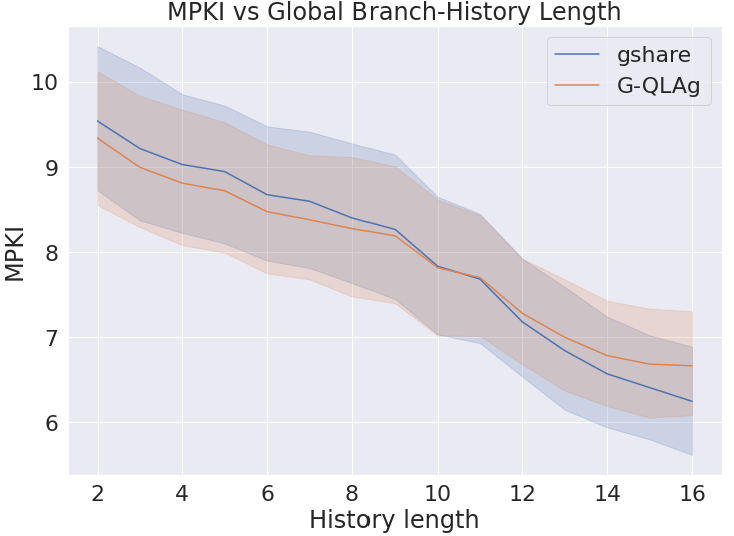}
\end{center}
\vspace{-4mm}
\caption{MPKI of gshare vs \GQlacron{} over all CBP-5 traces~\cite{cbp2016} as a function of global branch-history length. The error bands are proportional to the standard deviation over all traces.}
\label{fig:gshare}
\end{figure}

%
We have implemented in the CBP-5~\cite{cbp2016} framework the aforementioned $Q$-Learning based variant of gshare, which we call \emph{\GQlacron{}} (\GQlmethod{}). \GQlacron{} consists of a $N$-entry table, each containing two $Q$-values: $Q$\textsubscript{\TAKEN} and $Q$\textsubscript{\NOTTAKEN}. 
$Q$-values are signed floating-point numbers ranging from $-1$ to $1$. Experimentally, we found that using 6-bit $Q$-values is a good trade-off in our setup. Like gshare, \GQlacron{} is indexed with a hash of the branch PC and GHR modulo the size of the prediction table. For the matching entry, when $Q\textsubscript{\TAKEN} > Q\textsubscript{\NOTTAKEN}$, the prediction is \emph{\TAKEN}. Similarly, when $Q\textsubscript{\TAKEN} < Q\textsubscript{\NOTTAKEN}$, the prediction is \emph{\NOTTAKEN}. In the case where $Q\textsubscript{\TAKEN} = Q\textsubscript{\NOTTAKEN}$, the predicted direction (\emph{\TAKEN, \NOTTAKEN}) is picked at random with equal probability. $Q$-values are initialized to $0$. At update time, the reward $r$ is set to $1$ or $-1$ if the prediction was correct or incorrect, respectively. The $Q$-value that corresponds to the prediction ($\max(Q\textsubscript{\TAKEN},Q\textsubscript{\NOTTAKEN})$) is updated according to the formula in the previous paragraph with learning rate $\alpha=0.2$.
Figure~\ref{fig:gshare} compares the mispredictions per kilo-instructions (MPKI) of gshare and \GQlacron{} by varying the global branch-history length. We consider the same storage budget (64KB) for both predictors and configure their number of entries accordingly. Our implementation of \GQlacron{} uses 12 bits (2 $Q$-values) per entry, resulting in significantly fewer entries than gshare, which requires only 2 bits per entry.
%

%
As the figure shows, the two predictors perform similarly across the range of histories. As expected, for both predictors, MPKI reduces as history length is increased. However, at longer histories, the capacity disadvantage of \GQlacron{} becomes more pronounced, resulting in higher MPKI than gshare. 
%

%
As the study shows, it is indeed straightforward to cast existing branch predictors as RL, and can be done without considering RL aspects such as $\gamma$ and $\varepsilon = 0$ for \GQlacron{}. However, as we show in section~\ref{sec:rl-bp-methods}, considering these and other aspects of RL opens new venues for BP policy optimization.
%
%
%

%
%

%
\section{RL-based Branch Predictors: looking ahead} \label{sec:rl-looking-ahead}
%
%
\subsection{Exploring the BP Design Space through RL Methods}
\label{sec:rl-bp-methods}

%
\begin{table*}[ht]
\vspace{-5mm}
\begin{center}
  \begin{tabular}{|p{4cm}||p{1.75cm}|p{5cm}|p{2.25cm}|p{3.5cm}|}
    \toprule
    Branch Predictor  &  State space ($\states$) & Policy ($Q$-values / $\pi(a|s)$)  	&  Objective / Loss & Online Policy Optimizer \\    
    \midrule
    Bimodal~\cite{smith1981}             &  PC  & $2$-bit saturating counter & N/A & Approx. Q-Learning update \\
    \hline
     Gshare~\cite{mcfarling1993}             &  PC $\oplus$ GHR  & $2$-bit saturating counter & N/A & Approx. Q-Learning update \\
    \hline
     Perceptron~\cite{perceptron2001}             &  PC, GHR  & $\mathbf{sign}(b+w_1 ghr_1+\dots + w_l ghr_l)$ & Hinge Loss & Perceptron update \\
    \hline
    Hashed Perceptron (O-GEHL~\cite{ogehl_seznec2005})         & PC, GHR & As in Perceptron but hashed PC, GHR & Hinge Loss 	& Perceptron update   \\
    \hline
    Multiperspective Perceptron~\cite{jimenez2016}    & Many & Various branch features (ghist, pathhist, etc) & Hinge Loss 	& Perceptron update \& heuristics   \\
    \hline
    BranchNet~\cite{branchnet}    & PC, GHR 
    & Convolutional Neural Network (CNN) 
    & Logistic Loss
    & \textbf{Offline} Adam optimizer~\cite{rl:opt:adam}
    \\
    \hline
    \textbf{\Pgacron{}}   & PC, GHR & $\sigma(b+w_1 ghr_1+\dots + w_l ghr_l)$  & Equation~\ref{eqn:pg_objective} & REINFORCE (Section~\ref{sec:rl:policy-gradient-intro}) \\
    \bottomrule
  \end{tabular}
   \end{center}
  \caption{Branch predictors viewed from an RL perspective. Categorization is based on state representation, policy function, loss function and online policy optimization. $\sigma(x):=1/(1+\exp(-x))$ is the sigmoid function.} \label{tab:bp:objectives}
\end{table*}
\vspace*{-1mm}
%
Following the Policy-Gradient framework explained in Section~\ref{sec:rl:policy-gradient-intro}, we next discuss how RL can facilitate the development of future BPs. We identify four key ingredients that a branch predictor needs to specify: (1) policy, (2) state representation, (3) loss function optimized by the predictor and (4) optimization strategy. 
\vspace{1.5mm}
\noindent
\emph{Policy:} For branch prediction, the policy $\pi(a|s)$ is a function from the current state to a branch outcome. The policy may be expressed as a linear model or a non-linear model. 
Linear models are simple and might be readily amenable to a hardware-friendly implementation. The drawback of linear models is that they are unable to capture non-linear correlations, such as an XOR relationship. The Perceptron BP is an example of a design that employs a linear model and, hence, fails to capture correlations between non-linearly separable branches~\cite{perceptron2001}. 

Non-linear models, including multi-layer neural networks, are fundamentally more expressive than linear ones and have the capacity to capture non-linear correlations. The limitation of non-linear models, however, is that they do not land themselves to a straight-forward microarchitectural implementation. For instance, the recent BranchNet predictor~\cite{branchnet} attempts to overcome implementation limitations with a specialized software-hardware co-design: a plurality of convolutional neural networks (CNNs) are trained at compile-time (for up to 18 hours depending on the application) for a few hard-to-predict (H2P) branches. BranchNet's inference engine is implemented in hardware for predicting only those branches, alongside TAGE. While the design is shown to be effective, it is complex. We believe that finding the right balance between model simplicity and its capacity will be a major focus in future ML-inspired microarchitectural research.
%

%
\vspace{1.5mm}
\noindent
\emph{State representation:} In the context of branch prediction, the state may encompass branch addresses, global and/or local history bits, loop counters, etc. 
In Section~\ref{sec:rl-basics}, we briefly presented examples of Bimodal and gshare BPs, which can be viewed as instances of $Q$-Learning with PC and PC$\oplus GHR$ state representation, respectively.
\vspace{1.5mm}
\noindent
\emph{Loss function:} Also known as an objective function, this governs the goal that the policy seeks to achieve.
For a branch predictor, this should be minimizing the number of mispredictions. For instance, the Perceptron BP minimizes the number of mispredictions explicitly using the hinge loss function\footnote{Hinge loss for a binary classifier with output $y$ and target $t\in\{\pm 1\}$ is defined as $\max(0, 1-ty)$.}. However, the goal of a branch predictor does not have to \emph{explicitly} minimize the mistakes, but it could do so \emph{implicitly}. That is, an alternative goal is to minimize the \emph{misprediction probability} for a branch predictor that is probabilistic, i.e., returns the probability that the branch will be taken. Such a branch predictor optimizes the probability of predicting the correct branch direction and thus implicitly minimizes mispredictions. In Section~\ref{sec:PolGAg}, we will discuss the design and performance of such a predictor. 
\vspace{1.5mm}
\noindent
\emph{Optimization strategy:} In RL, the optimization strategy consists of an \emph{optimizer}, which is a method to minimize the loss function, along with additional ``knobs" 
that, e.g., control the learning rate. In the BP domain,  the optimization strategy defines the branch predictor update rule. For example, the Perceptron BP uses the Perceptron update rule, which is implicitly an online gradient descent optimizer with hinge loss (see Table~\ref{tab:bp:objectives}). In the O-GEHL predictor~\cite{ogehl_seznec2005}, the dynamic threshold that guides the predictor's update can be roughly thought as an adaptive learning-rate adjustment for the optimizer.

%
Table~\ref{tab:bp:objectives} summarizes these design elements for a selection of well-known branch predictors, as well as \Pgacron{}, a Policy-Gradient based branch predictor introduced in the next section.
%
\vspace{-5mm}
\subsection{Example: A Policy-Gradient Based Branch Predictor} \label{sec:PolGAg}
%
In this section, we study the Perceptron BP through the perspective of Policy Gradient, with the aim of showcasing how the RL construction can be used to explore the BP design-space.

%

%

%
%
%
To that end, we design a Perceptron-like branch predictor, which we call \emph{\Pgmethod{} BP} (\Pgacron{}). The \Pgmethod{} is based on the REINFORCE algorithm~\cite{book:Sutton1998}, a well-known Policy-Gradient method. Specifically, we adjust the REINFORCE algorithm for the context of the Perceptron BP. The resulting design is described with Algorithm~\ref{alg:linear_policy_gradient}. We derive Algorithm~\ref{alg:linear_policy_gradient} from a holistic optimization view of the problem, i.e., maximizing the objective function of Equation~\ref{eqn:pg_objective} and making specific decisions on the definition of the policy $\pi_{\vectheta}(a|s)$. 
%
\vspace{-1mm}
\begin{algorithm}[h]
	\caption{\Pgmethod{} BP (\Pgacron{})}\label{alg:linear_policy_gradient}
\begin{algorithmic}[1]
\Procedure{PolGAg}{$l$, $\alpha$}\Comment{$l$ GHR bits/learning rate $\alpha$.}
\State Set $\vectheta[PC] = \zero$ for every PC.
\For {each branch PC}
\State Let $a\in\{\TAKEN, \NOTTAKEN \}$ with reward $r\in{\{\pm 1\}}$.
\State $\vectheta[PC] \mathrel{{+}{=}} \alpha r \pi_{\vectheta}(\bar{a}|s) \x(s,a)$.
\EndFor
\State \textbf{Output:} Branch prediction policy: $\pi_{\vectheta}(a|s)$.
\EndProcedure
\end{algorithmic}
\end{algorithm}
\vspace{-1mm}
%
%

%
\Pgacron{} represents a specific design point with a policy, state space, loss function and optimization strategy (Section~\ref{sec:rl-bp-methods}). 
It suffices to define the policy and the state space $\states$, since the REINFORCE method already specifies the loss function (Equation~\ref{eqn:pg_objective}) and the optimization strategy (online gradient ascent).
For the policy, we use a linear softmax policy function with $h(s,a,\vectheta)=\vectheta^\top \x(s,a)$. Recall that $\x(s,a)$ is a vector representation of the state-action space. The state space $\states$ is defined as the set of all branch addresses (PC) and the content of GHR with length $l$. GHR is viewed as a vector $\q\in{\{\pm 1\}}^l$. For the purpose of the analysis, assume that there are $p$ unique branch addresses during the whole program execution. Also, let $\text{OHE}(PC)\in\{0,1\}^p$ be the one-hot encoding\footnote{In the one-hot-encoding representation, each coordinate corresponds to a unique address. $\text{OHE}(PC)$ has $1$ in the coordinate that corresponds to PC and $0$, otherwise.} (OHE) representation of the branch address. Moreover, define
$\q(\TAKEN):= [1, \q]$ and $\q(\NOTTAKEN) = -\q(\TAKEN)$, where the constant $1$ corresponds to a bias term. Finally, define $\x(s,a)$ as the $(p(l+1))$-vector 
$\x(s,a) = \text{OHE}(PC) \otimes \q $, where $\otimes$ denotes the Kronecker product. $\x(s,a)$ is defined for every state and action.
%
%
Using the above policy and state representation, the REINFORCE method translates to a specific update step of our \Pgacron{} Algorithm. 
Continuing from Section~\ref{sec:rl:policy-gradient-intro}, a direct calculation shows that $\nabla J(\vectheta) = \x(s,a)-\bm{\mu}=2\pi(\bar{a}|s) \x(s,a)$, where $\bar{a}$ flips the action from \TAKEN{}  to \NOTTAKEN{} or vice-versa. Therefore, Step 5 of Algorithm~\ref{alg:linear_policy_gradient} becomes $\vectheta \leftarrow \vectheta + 2 \alpha r \pi(\bar{a}|s) \x(s,a)$. 
%
%
\begin{figure}
\begin{center}
   \includegraphics[width=1.0\linewidth]{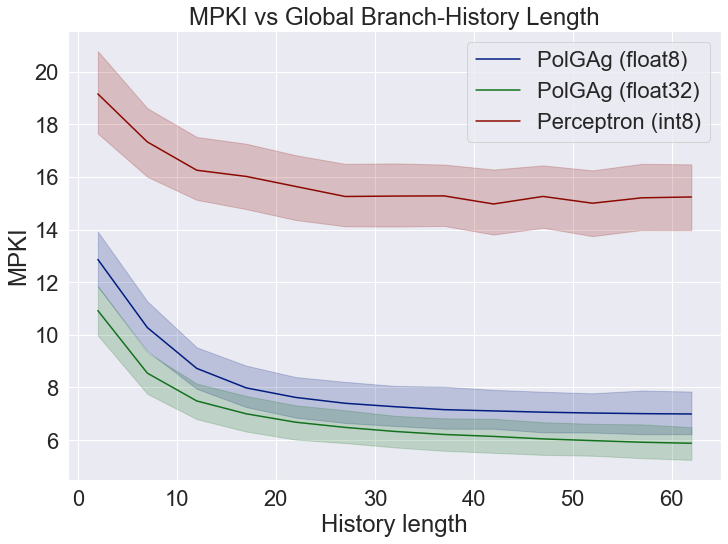}
\end{center}
\vspace{-5mm}
\caption{MPKI for \Pgacron{} and Perceptron predictors as a function of history length over all CBP-5 traces. Learning rate parameter $\alpha$ is $0.01$.}
\vspace{-5mm}
\label{fig:pg_vs_perceptron}
\end{figure}
%
%
%

%
In Figure~\ref{fig:pg_vs_perceptron}, we compare the MPKI of Perceptron~\cite{perceptron2001} and \Pgacron{} predictors. To avoid aliasing effects, we consider both predictors with unbounded storage, i.e., each branch has a dedicated set of weights. Firstly, for both predictors, each weight entry is stored using 8-bits. For \Pgacron{}, we use 8-bit floats  (1-bit sign, 5-bit exponent, 2-bit Mantissa) and for Perceptron 8-bit integers. As such, \Pgacron{} has more precision by employing floats, but also a higher computation complexity. However, as this is a limit study, we ignore the higher computational requirements of \Pgacron{}. 
As the figure shows, \Pgacron{} has a much higher prediction accuracy than the Perceptron. We attribute the difference to two factors: (a) the use of floats for \Pgacron{} and (b) the use of logistic loss and policy gradient update versus Perceptron's hinge loss update. 
Unfortunately, it is not easy to quantify the effect of each of these factors separately, since the choice of the loss function and update rule is coupled. In terms of using floats vs ints, the update rule of \Pgacron{} involves multiplication with a prediction probability, which makes it unfriendly to an integer representation. Instead, we assess a variant of \Pgacron{} that uses float32, rather than float8, for each weight. We observe that the use of higher precision further improves \Pgacron{}'s accuracy, reducing MPKI by a full point compared to \Pgacron{}-float8, which leads us to conclude that numerical precision does play a role in predictor's efficacy.
%

%
To summarize, we showed that the Perceptron predictor can be viewed as a Policy-Gradient method by appropriate selection of policy, state space, loss function and optimization strategy. While the resulting predictor, \Pgacron{}, was shown to outperform the baseline Perceptron, the former may be implementation-unfriendly due to its use of floats. 
Rather than espousing one design or the other, our purpose is to highlight the fact that viewing branch predictors through the RL prism opens new optimization opportunities.
%
\vspace{-1mm}
\subsection{Additional Examples}\label{sec:additional_examples}
%
We briefly show two more branch predictors that can be viewed in the Policy-Gradient framework: 

%
\vspace{1.5mm}
\noindent
{\bf O-GEHL~\cite{ogehl_seznec2005}.}
%
Here, we demonstrate that the O-GEHL BP is an instance of the Policy Gradient framework. The parameter space $\vectheta$ of O-GEHL is the concatenation of the $8$ tables each of which have roughly $8K$ $8$-bit entries. For any branch address (PC) and GHR instance, $\x (s,a)$ has only $8$ non-zero entries: one per table. The policy optimizer is the Perceptron update rule, the loss function is hinge loss.

%
\vspace{1.5mm}
\noindent
{\bf Multiperspective Perceptron~\cite{jimenez2016}.}
%
This latest Perceptron variant can also be viewed in the policy optimization setting. The only difference with our previous example is the state representation (more data must be collected to be able to compute the features) and the need of a more elaborate feature engineering, i.e. definition of the function $\x(s,\alpha)$.
%
\vspace{-1mm}
\subsection{Speculation \& RL Exploration}\label{sec:rl:exploration}
%
%
%
%
RL methods allow the agent to explore unknown states, potentially by making a decision likely to be disadvantageous in the short term, in order to achieve a higher cumulative long-term reward. We posit that such an approach could benefit existing or new BP schemes. In the RL-based BP paradigm, the exploration process would effectively require the agent (predictor) to consider information from wrong execution paths. But how to explore a wrong execution path without greatly affecting processor performance? 

We observe that on a misprediction, the BP continues to make predictions (and, accordingly, advance processor's control flow) up to the point of the pipeline flush. In today's BP designs, whatever was learned by the BP on the mispredicted path is discarded. However, the path observed under a misprediction may contain information valuable in later phases of the execution. In RL terms, there exists an opportunity to improve prediction accuracy through exploration and learning under a misprediction. In the case of $Q$-Learning, for example, to fully apply its update rule (Sec.~\ref{sec:rl:tabular}), it would be necessary to consider in the predictor's update the first branch address on the wrong path. In this way, the agent would be able to perform \emph{exploration} of the control-flow path according to the definition of the $\varepsilon$-greedy update step. To the best of our knowledge, the only proposed branch predictor that considers a similar option is the prophet/critic hybrid scheme~\cite{falcon2004}. We believe that the RL formulation opens the door to more systematic studies of exploration.

\vspace{-2mm}
\section{Conclusion \& Future Work}
%
This paper makes the case for \emph{Reinforcement Learning} based branch prediction as a new way of systematically developing future BPs. Our first-stage exploration demonstrates how RL concepts map naturally to the BP principles, how existing BP designs can be cast as RL models, and how such a formulation can open untapped opportunities. Our future work will focus on further exploring the opportunities offered by the RL formulation, and on converting these opportunities into hardware-friendly implementations.



\bibliographystyle{IEEEtranS}
\bibliography{main}
%
%
\section{Appendix: Exercising Policy Design in a Reinforcement learning gym}\label{appendix:gym}
%
RL methods adhere to a well-defined interface with any possible environment (the agent \textbf{acts} on the environment which in turn \textbf{steps} to the next state after providing a reward on the chosen action). Based on this interface, an open-source framework called Gym~\cite{rl:gym} has been developed to facilitate research of new RL algorithmic approaches on various environments, such as bipedal robot walking (Humanoid-v2) and playing a specific video game (Pong-ram-v0) to name a few. As a major benefit, Gym allows the implementation of custom environments for new problems where RL methods could be found useful to address them. Similarly, we have extended\footnote{The tool is available to download on the workshop's webpage: \url{https://sites.google.com/view/mlarchsys/isca-2021}.} Gym by defining several custom BP environments (\gymBP) based on the public traces of CBP-5 to explore new RL policies on branch prediction.
Each of these environments is defined for a specific branch instruction of a given trace from all the $440$ traces of CBP-5. For instance, the code snippet below initializes an environment for branch with PC address equal to $65872$ of \emph{SHORT\_MOBILE-42} trace.
%

%
\vspace{0.35cm}
\begin{minipage}{.45\textwidth}
\begin{lstlisting}[frame=tb,language=python]
import pandas as pd
import gym
import gym_bp_cbp2016

# Set trace, branch and global history length 
df = pd.read_parquet("SHORT_MOBILE-42.parquet")
branch = 65872
ghr_len = 17

# Initialize gym_bp_cbp2016 environment
env = gym.make('bp-cbp2016-v0')
env.init(df,ghr_len,branch)
\end{lstlisting}
\end{minipage}%
%

%
An RL-based BP interacts with the \gymBP{} environment via the step method as listed below. The branch predictor inputs the branch direction (T/NT) as an integer type with 0/1 values and the environment responds with the next observation (GHR bits), the reward ($\pm 1$) and a boolean flag `done` if the environment has reached the end. The RL-based BP now has access to observations and rewards and it can update its policy and make the next prediction.
%

%
\vspace{0.35cm}
\begin{minipage}{.45\textwidth}
\begin{lstlisting}[frame=tb,language=python]
# Predict branch direction and environment step
pred_branch_dir = agent.act(previous_obs) 
obs,reward,done,_ = env.step(pred_branch_dir)
\end{lstlisting}
\end{minipage}%
%

%
The above code snippet shows a branch prediction step followed by a next state step of the environment.

\end{document}